\documentclass[10pt,twocolumn,letterpaper]{article}

\usepackage[pagenumbers]{cvpr} 

\usepackage{graphicx}
\usepackage{amsmath}
\usepackage{amssymb}
\usepackage{booktabs}
\usepackage{multirow}

\usepackage[pagebackref,breaklinks,colorlinks]{hyperref}

\usepackage[capitalize]{cleveref}
\crefname{section}{Sec.}{Secs.}
\Crefname{section}{Section}{Sections}
\Crefname{table}{Table}{Tables}
\crefname{table}{Tab.}{Tabs.}

\def\cvprPaperID{*****} 
\def\confName{CVPR}
\def\confYear{2022}

\begin{document}
	
	\title{Toward Accurate and Reliable Iris Segmentation Using Uncertainty Learning}
	\author{
		Jianze Wei, 
		Huaibo Huang, 
		Muyi Sun, 
		Yunlong Wang,
		Min Ren,
		Ran He, 
		Zhenan Sun\\
		School of Artificial Intelligence, University of Chinese Academy of Sciences, Beijing, China\\
		CRIPAC \& NLPR, Institute of Automation, Chinese Academy of Sciences, Beijing, China\\
		{\tt\small  \{jianze.wei, huaibo.huang, muyi.sun, yunlong.wang, min.ren\}@cripac.ia.ac.cn}\\ {\tt\small\{rhe, znsun\}@nlpr.ia.ac.cn}
	}
	\maketitle
	
	\begin{abstract}
		Iris segmentation is a deterministic part of the iris recognition system. 
		Unreliable segmentation of iris regions especially the limbic area is still the bottleneck problem, which impedes more accurate recognition. 
		To make further efforts on accurate and reliable iris segmentation, we propose a bilateral self-attention module and design \textbf{Bi}lateral \textbf{Trans}former (BiTrans) with hierarchical architecture by exploring spatial and visual relationships.
		The bilateral self-attention module adopts a spatial branch to capture spatial contextual information without resolution reduction and a visual branch with a large receptive field to extract the visual contextual features.
		BiTrans actively applies convolutional projections and cross-attention to improve spatial perception and hierarchical feature fusion.
		Besides, Iris Segmentation Uncertainty Learning is developed to learn the uncertainty map according to prediction discrepancy.
		With the estimated uncertainty, a weighting scheme and a regularization term are designed to reduce predictive uncertainty.
		More importantly, the uncertainty estimate reflects the reliability of the segmentation predictions.
		Experimental results on three publicly available databases demonstrate that the proposed approach achieves better segmentation performance using 20\% FLOPs of the SOTA IrisParseNet.
		
	\end{abstract}

	\section{Introduction}\label{sec_intro}
	With the increasing demands of biometrics available in the non-invasive and masked face scenario, researchers gradually consider achieving personal identification using ocular biometrics, including iris recognition known for its high accuracy.
	Iris pattern builds its uniqueness through the rich texture details (such as freckles, coronas, crypts, furrows)~\cite{sun2008ordinal,he2008toward}, which establishes a solid cornerstone in the high accuracy of iris recognition.
	To find the discriminative clue for recognition, iris segmentation is proposed to segment the valid region that contains rich iris texture out from the ocular image.
	
	However, there are multiple ocular components in the captured ocular image, including iris, pupil, sclera, eyelashes, eyelids, eyebrows, etc. 
	These components have not only special spatial layouts but also their own visual characteristics. 
	For example, the pupil is black and located in the center of the human eye.
	Iris segmentation separates the iris from ocular components and labels the rest components as non-iris regions.
	This one-vs-rest classification task encourages segmentation models to explore the contextual relationship from the spatial and visual perspectives. 
	
	To model the contextual relationship, some pioneering works are proposed based on extra prior knowledge, such as circular boundary assumption~\cite{daugman1993high, wildes1997iris}, total variance~\cite{zhao2015accurate}, low-resolution contour~\cite{sutra2012viterbi}.
	With the development of deep iris segmentation, researchers realize that it is feasible to model the context by exploring the inner knowledge in the images.
	MFCNs~\cite{liu2016accurate} investigates the contextual information at multiple spatial scales and fuses the multi-scale features for segmentation prediction.
	FCDNN~\cite{bazrafkan2018end} explores the context in the noise scene and proposes an end-to-end solution for noisy iris segmentation.
	To learn the long-range relationship, IrisParseNet~\cite{wang2020towards} proposes a multi-scale attention module with three dilated convolutions to extract multi-scale context features.
	These methods gradually reveal that intrinsic knowledge for iris segmentation is hidden in the contextual relationship.
	However, their local convolutional kernels limit the context modeling for long-range dependencies.
	Recently, the visual transformer has developed dramatically and achieved remarkable advances in semantic segmentation task~\cite{liu2021Swin, zheng2021rethinking}.
	Different from previous CNN models, transformers employ a self-attention mechanism to explore long-range dependencies in the image and mitigate the limited context modeling problem.

	Inspired by the above analysis, the paper proposes a transformer-style backbone named \textbf{Bi}lateral \textbf{Trans}former (BiTrans) with a residual encoder-decoder architecture for iris segmentation. 
	BiTrans leverages the transformer's powerful capability in modeling long-range relationships to explore the visual and spatial characteristics.
	In BiTrans, we adopt a bilateral self-attention module with spatial and visual branches for contextual aggregation.
	The former extracts spatial contextual information while preserving feature resolution.
	The latter applies a large receptive field to find the discriminative contextual clues about visual characteristics.
	The bilateral self-attention module builds a comprehensive relationship by integrating contextual information from spatial and visual branches.
	Besides, the transformer block of BiTrans (i.e., BiTrans block) makes two other efforts for iris segmentation.
	First, it utilizes convolutional projections to replace linear projections that are not sensitive to spatial position, improving BiTrans's spatial perception.
	Second, a cross-attention version is developed to build a symmetric structure for the fusion of hierarchical features.
	In the experiments on three iris segmentation, the proposed BiTrans shows a competitive performance with the SOTA IrisParseNet in iris segmentation.

	In addition, we propose \textbf{I}ris \textbf{S}egmentation \textbf{U}ncertainty \textbf{L}earning (ISUL) for better segmentation reliability.
	As the primary process of iris recognition systems, the segmentation reliability determines the system's trustworthiness.
	The deep iris segmentation's reliability is mainly driven by two aspects, annotation noise and model uncertainty.
	The former is highly correlated to the subjective feelings of annotators, while the latter is an inevitable challenge for deep learning.
	To mitigate the problem, ISUL only introduces an auxiliary head and embeds it to the hidden layers to predict an additional segmentation result, as shown in Figure~\ref{fig_framework}.
	Since layers at different depths learn distinct segmentation features~\cite{antoran20depth}, ISUL leverages this discrepancy to estimate the uncertainty map.
	The uncertainty map reflects the reliability of the segmentation results.
	Considering that the auxiliary head is usually applied for deep supervision~\cite{lee2015deeply}, the proposed ISUL can be regarded as a low-cost module for uncertainty estimation.
	Based on the uncertainty map, a weighting scheme and a regularization term are developed to improve the accuracy and reliability of the iris segmentation model.
	The weighting scheme assigns large values to the pixels with high uncertainty, hence, large gradients would flow through these pixels in the backward propagation.
	As for the regularization term, the uncertainty map can also work as a part of it to further reduce the prediction uncertainty.


	The main contributions are summarized as follows:
	\begin{itemize}
		\item We propose a bilateral self-attention module to capture the visual and spatial relationships for iris segmentation.
		In the module, the spatial branch extracts spatial dependencies without resolution reduction, while the visual branch captures visual context using a large receptive field.
		Based on the module, we design BiTrans with a residual encoder-decoder network for iris segmentation.	
		\item	ISUL is proposed to learn the predictive uncertainty by only introducing an auxiliary head.
		Based on estimated uncertainty, a weighting scheme and a regularizer are developed to improve the model reliability.	
		\item The superior performance on three challenging iris datasets demonstrates the effectiveness of the proposed approach.
		In addition, a comprehensive ablation study is conducted to understand the contributions of each module.
		
	\end{itemize}

	
	\section{Related Work}
	
	\subsection{Iris Segmentation}
	A complete iris recognition consists of four parts, image acquisition, pre-processing, feature extraction, and matching.
	Iris segmentation is an important step of pre-processing, and it segments out the valid iris region for subsequent tasks.
	
	Conventional iris segmentation approaches pursue accurate iris segmentation by using extra prior knowledge or assumption.
	One of the most famous assumptions is the circular assumption~\cite{daugman1993high, wildes1997iris}, i.e.,  the inner and outer boundaries of the iris region can be regarded as the perimeters of two parameterized circulars.
	Since circle parameters are also the goal of the iris localization task, it is hard to tell apart iris segmentation and localization clearly.
	Integro-differential operator~\cite{daugman1993high} is adopted to search the satisfying circle matching the maximum variety of image intensity.
	Circle Hough Transform~\cite{wildes1997iris} searches for the circle parameters according to edge points that fall on perimeters of circles.
	The above two segmentation methods are extended to application scenarios with fewer constraints by introducing extra knowledge, such as total variation~\cite{zhao2015accurate}, low-resolution contour~\cite{sutra2012viterbi}, pull and push model~\cite{he2008toward}.
	Besides the circular assumption, the hypothesis of curve shapes is also introduced and applied in some works~\cite{banerjee2015iris,shah2009iris} to fit inner and outer boundaries.
	
	
	\begin{figure*}[ht]
		\centering
		\vspace{-1mm}
		\includegraphics[width=0.9\textwidth]{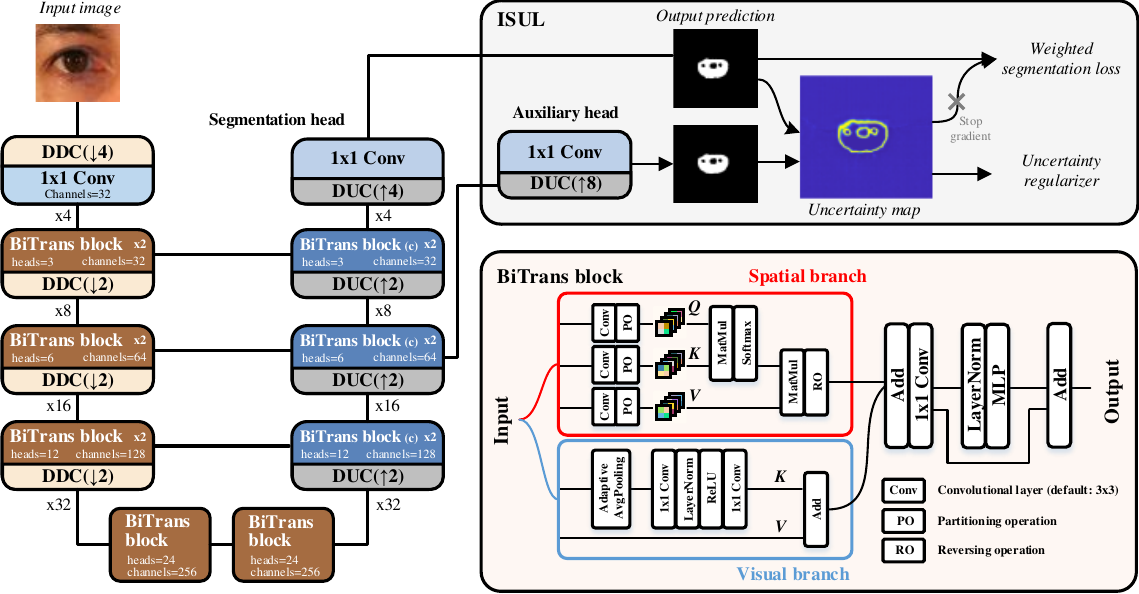}
		\vspace{-2.5mm}
		\caption{The framework of the proposed approach.
			The BiTrans block adopts the bilateral self-attention module with two parallel branches to capture visual and spatial contextual information. 
			BiTrans block (c) represents the cross-attention version.
			ISUL only introduces an auxiliary head to generate the additional segmentation prediction, and utilizes the difference between predictions to estimate uncertainty.
		}
		\vspace{-4mm}
		\label{fig_framework}
	\end{figure*}
	
	Recently, the impressive progress of deep models in semantic segmentation has promoted the development of deep iris segmentation.
	Deep iris segmentation does not rely on external knowledge but explores the intrinsic context in the images, which makes the boundary between iris segmentation and localization clearer.
	MFCNs~\cite{liu2016accurate} pioneers the application of FCN (Fully convolutional networks) to iris segmentation. 
	In addition to FCN, encoder-decoder models~\cite{arsalan2019fred} and U-Net models~\cite{lozej2018end, lian2018attention} are also widely used in iris segmentation.
	CNNHT~\cite{hofbauer2019exploiting} introduces circular assumption and applies circular hough transforms to generate iris boundaries.
	IrisParseNet~\cite{wang2020towards} develop the U-Net and proposes a multi-task attention network to learn iris mask, iris outer boundary, and
	pupil mask, simultaneously.

	\subsection{Visual Transformer}
	In recent years, the transformer has made impressive progress in multiple research fields, its rapid development attracts more and more attention from natural language processing (NLP) and computer vision (CV).
	Transformer models the long-range dependencies based on self-attention and builds a new architecture that is significantly different from the CNN and RNN.
	Vaswani et al.~\cite{vaswani2017attention} first propose a transformer for a NLP task and define its initial structure.
	With the proposition of transformer, many influential works emerge in multiple NLP tasks, such as BERT~\cite{devlin2018bert},
	GPT2~\cite{radford2019language},
	BigBird~\cite{zaheer2020big}.
	
	The success of the transformer model in NLP quickly caught the eyes of the CV community.
	People start leveraging the transformer to explore the data context for the CV tasks.
	Initially, some approaches~\cite{bello2019attention,sun2019videobert} develop existing CNN models by introducing partial modules of the transformer model, like self-attention.
	Their exciting experimental results encourage researchers to adopt a pure transformer model for the various CV tasks.
	In classification, ViT~\cite{dosovitskiy2020image} adopts patch embedding to generate sequential features for the image and builds a whole classification model using a pure transformer.
	For object detection, DETR~\cite{carion2020end} designs a new object detector based on transformers and discards the need for some hand-designed components like anchor boxes and Non-Maximum Suppression (NMS).
	For image generation, TransGAN~\cite{jiang2021transgan} leverages the power of the transformer and carefully designs a GAN model without convolutional operations.
	For semantic segmentation that is closely related to iris segmentation, there are a lot of methods based on the transformer-style architecture.
	SETR~\cite{zheng2021rethinking} designs a pure transformer model for segmentation and models global context in every layer of the transformer.
	TransUnet~\cite{chen2021transunet} inserts a transformer module into U-Net to improve the feature generation of the bottleneck network.
	Swin transformer~\cite{liu2021Swin}  proposes a shift window strategy to generate hierarchical feature maps.
	
	
	\section{Method}
	This section will introduce the technical details of the proposed BiTrans and ISUL.
	
	\subsection{Architecture Overview}
	The overall architecture of the proposed method is illustrated in Figure~\ref{fig_framework}.
	The proposed method adopts a residual encoder-decoder architecture consisting of encoder, bottleneck, decoder, skip connections, segmentation head, and auxiliary head.
	In BiTrans, the basic unit for contextual aggregation is the BiTrans block.
	The encoder and bottleneck adopt the vanilla BiTrans block, while the decoder adopts its cross-attention version.
	
	The encoder is alternately stacked by Dense Down-sampling Convolutions (DDCs) and BiTrans blocks.
	DDC is the down-sampling version of the Dense up-sampling Convolution (DUC), and it is used to reduce the input resolution for multi-scale feature generation.
	DDC divides the iris image into a grid of non-overlapping patches and arranges the pixels in the patch to different channels.
	Then, an $1\times1$ convolutional layer is placed at the end of DDC to project arranged features to certain channels.
	More details about DDC are displayed in Appendix-A.
	DDC generates features with different spatial resolutions while BiTrans blocks model the spatial and visual context of these features.
	In the forward propagation stage, an ocular image travels through the combination of DDCs and BiTrans blocks to produce hierarchical feature representations.

	As for the decoder, we design a symmetrical structure with the encoder.
	Specifically, BiTrans blocks and DUCs construct the main body of the decoder.
	These DUCs are inserted into the decoder for an opposite goal of DDCs, i.e., increasing the spatial resolution of features.
	The BiTrans block leverages the cross-attention to guide the generation of segmentation prediction according to multi-scale features from skip connections.
	
	In the end, a $4\times$ DUC and an $1\times 1$ convolutional layer constitute the segmentation head.
	The segmentation head predicts a pixel-wise segmentation result via up-sampling.
	In addition, we adopt an auxiliary head for ISUL. 
	The auxiliary head has the same structure as the segmentation head, but it employs an $8\times$ DUC rather than a $4\times$ module to restore the spatial resolution.
	In the proposed approach, the auxiliary head is not only used for deep supervision~\cite{lee2015deeply} but also adopted for the uncertainty estimation.
	An auxiliary head is free in a sense if we do not demand the reliability of the model in the inference stage.
	
	In the subsequent parts, we will elaborate on BiTrans and ISUL.
	
	\subsection{Bilateral Transformer}
	As mentioned in Section~\ref{sec_intro},  iris segmentation is a one-vs-rest classification task, and there are multiple ocular components with special spatial and visual characteristics.
	To capture the spatial and visual characteristics, we propose a bilateral self-attention module and then elaborately design the BiTrans block.

	\subsubsection{Bilateral Self-Attention Module}
	Since different ocular components contain special spatial and visual characteristics, the bilateral self-attention module applies spatial and visual branches to learn contextual clues for two characteristics, respectively.
	
	\noindent{\textbf{Spatial Branch.}}
	Published literature~\cite{yu2018bisenet,chen2017deeplab} suggests that spatial information is crucial for accurate segmentation, and the resolution reduction will lose the majority of spatial information, leading to performance degeneration.
	In order to preserve the loss of spatial information, the spatial branch achieves contextual aggregation using the resolution of the input image.
	Considering that captured ocular images are high resolution due to iris quality standards like ISO-SC37, ISO-19794, ISO-29794, the spatial branch adopts a shifted window strategy~\cite{liu2021Swin} to reduce computational consumption without the resolution reduction.
	
	Specifically, given a feature map $X\in\mathbb{R}^{C\times H\times W}$, $H$ and $W$ are the height and width of the feature map, $C$ represents the number of the semantic classes,
	then we can obtain the attention inputs, including
	query matrix $Q=XW^Q$, 
	key matrix $K=XW^K$,
	and value matrix $V=XW^V$,
	where $W^Q$, $W^K$, $W^V$ represent the projection matrices for $Q$, $K$, $V$, respectively.
	We utilize partitioning operation ($\text{PO}$) to divide the attention inputs into $\frac{H}{M}\times\frac{W}{M}$ non-overlapping patches of the same size $M\times M$ and tokenize them.
	The self-attention for spatial branch (SB-MHSA) can be computed as follow, 
	\vspace{-2mm}
	\begin{equation}
	\vspace{-1mm}
	\begin{aligned}
	\text{SB-MHSA}(X)=&\text{Attention}(Q,K,V)\\
	=&\text{RO}(A_1,...,A_{N}),\\
	\end{aligned}
	\end{equation}
	where 
	$\text{RO}$ denotes the reverse operation to reverse the divided patches into a complete feature map with input resolution.
	$N=\frac{H}{M}\times\frac{W}{M}$ is the number of the non-overlapping patches.
	$A_i$ is the attention output of the $i$-th patch, and it can be computed by 
	\vspace{-1mm}
	\begin{equation}
	A_i=\text{SoftMax}(\frac{\text{PO}(Q)_i\text{PO}(K)_i^T}{\sqrt{d}}+B)\text{PO}(V)_i,
	\end{equation}
	where
	$\text{PO}(\cdot)_i$ denotes the $i$-th divided patch.
	$B$ is the $M\times M$ relative position bias~\cite{liu2021Swin}.

	\noindent{\textbf{Visual Branch.}}
	Since ocular components with different visual characteristics are located in different spatial positions, it is natural to apply the receptive field with a large size to capture visual relationships.
	Inspired by the finding~\cite{cao2019gcnet} that self-attention provides almost the same attention map for different query points, we develop the visual branch to learning visual contextual aggregation.
	
	Different from the spatial branch, the visual branch discards the query matrix ($Q$) and updates contextual features according to key and value matrices according to the mentioned finding.
	Specifically, the visual branch does not hesitate to reduce the spatial resolution of the attention input and learn visual clues using adaptive average pooling.
	Since the other branch has learned spatial information, it is feasible to reduce the spatial resolution in the visual branch.
	In addition, the smaller spatial resolution contributes to reducing computational consumption.
	The computational process for the visual branch (VB-MHSA) can be formulated as 
	\begin{equation}
	\begin{aligned}
	\text{VB-MHSA}(X)=&V+Q\\
	=&X+\text{VBNet}(X),\\
	\end{aligned}
	\end{equation}
	where VBNet denotes a visual-branch sub-network with the architecture of ``AdaptiveAvgPool-Conv-LayerNorm-ReLU-Conv'', as shown in Figure~\ref{fig_framework}.
	For the contextual features from two branch, the bilateral self-attention module fuses them in a summation manner, i.e., 
	\vspace{-2mm}
	\begin{equation}
	\vspace{-1mm}
	\text{Bi-MHSA}(X)=\text{SB-MHSA}(X) + \text{VB-MHSA}(X).
	\end{equation}

	\subsubsection{BiTrans Block}
	The BiTrans block is built upon the bilateral self-attention module, and it is also the basic aggregation unit of our segmentation model.
	In order to make the bilateral self-attention module work well in the hierarchical encoder-decoder network, we make the following two efforts.
	
	The first one is introducing convolutional projections to replace linear projections.
	In front of the self-attention module, the visual transformer adopts linear projections to generate self-attention inputs (i.e., $Q$, $K$, and $V$).
	The linear projections aggregate the pixels in the same window but do not care about where the pixels are.
	One of the most extreme cases is the position of the corner of the window, each position is close to eight neighboring pixels. 
	However, linear projection utilizes only three of them and other faraway pixels in the same window to generate attention inputs. 
	To further improve spatial perception, BiTrans block uses a position-sensitive operation, convolutional projections, to replace linear projections.
	Specifically, given feature maps $X\in\mathbb{R}^{C\times H\times W}$, the BiTrans block generates the attention inputs 
	$Q=\phi_Q(X)$, 
	$K=\phi_K(X)$,
	and
	$V=\phi_V(X)$,
	$\phi_Q$, $\phi_K$, and $\phi_V$ are the convolutional projections for attention inputs $Q$, $K$, $V$, respectively.

	The other effort is that we develop a cross-attention version of the BiTrans block for the decoder.
	Considering that semantic segmentation is a pixel-wise task, it is vital to utilize multi-scale features generated by the encoder.
	The cross-attention version of the BiTrans block is developed to better leverage hierarchical features.
	Different from the vanilla BiTrans block, the cross-attention version employs cross-attention modules to generate attention maps.
	Specifically, it utilizes the hierarchical features from skip connections to generate $Q$.
	The change in attention inputs turns self-attention into cross-attention and allows multi-scale visual features to join the generation of the attention map.
	In the decoder, the cross-attention fuses the hierarchical features to modify the attention map and aggregate contextual features,

	For two successive transformer blocks, the BiTrans block acts like Swin transformer blocks.
	The first one uses the regular window strategy, and the second one uses the shifting strategy (named SBi-MHSA).
	The continuous BiTrans blocks can be formulated as:
	\begin{equation}
	\begin{aligned}
	\hat{X}^l=&\text{Bi-MHSA}(\text{LN}(X^{l-1})) \\
	X^l=&\text{MLP}(\text{LN}(\hat{X}^l))+\hat{X}^l\\
	\hat{X}^{l+1}=&\text{SBi-MHSA}(\text{LN}(X^{l}))\\
	X^{l+1}=&\text{MLP}(\text{LN}(\hat{X}^{l+1}))+\hat{X}^{l+1}\\
	\end{aligned}
	\end{equation}
	where
	$\hat{X}^{l}$ and ${X}^{l}$ denote the outputs of the bilateral self-attention module (Bi-MHSA) and the Multi-Layer Perceptron (MLP) module of the $l$-th block.
	LN is short for Layer Normalization.
	
	\subsection{Iris Segmentation Uncertainty Learning}
	Due to annotation noise and model uncertainty, it is critical to learn the reliability of the segmentation predictions.
	However, the current technology usually estimates uncertainty at the expense of increased computational cost, such as multiple forward propagations or deep ensemble, which is unsuitable for iris segmentation that demands real-time efficiency and small model size.
	Here, we propose ISUL for uncertainty estimation and apply it to a weighting scheme and a regularization term for accurate and reliable results.
	
	\subsubsection{Uncertainty Estimation}
	For a neural network, different depths correspond to distinct subnetworks~\cite{antoran20depth}, and they make different predictions for the same input data.
	The discrepancy between predictions reflects the uncertainty/reliability of the predictions.
	Motivated by this intuitive idea, we propose ISUL to estimate uncertainty with a single forward propagation by only introducing an auxiliary head.

	Assuming that $P_s,P_a\in\mathbb{R}^{H\times W\times C}$ are the predictions from the segmentation head and auxiliary head, respectively, then we can obtain a pixel-wise uncertainty map $M\in\mathbb{R}^{H\times W}$ according to the discrepancy of predictions.
	The computation process
	can be written as 
	\begin{equation}\label{eq_reliable}
	M=-\sum_{c}\left(P_s\log{P'} +P_a\log{P'}\right),
	\end{equation}
	where $P'=P_s+0.5\times P_a$ denotes the average segmentation prediction.
	$c\in \{1,..., C\}$ represents the $c$-th semantic label, and $C=2$ in iris segmentation.

	Compared with the previous method, the proposed approach only adds an auxiliary head and utilizes the predictions from the last two layers for the uncertainty estimation.
	In addition, the auxiliary head usually works as the module of deep supervison~\cite{lee2015deeply} in the training stage to further improve segmentation accuracy.
	In other words, we obtain an uncertainty estimate at a low cost.

	\subsubsection{Weighting Scheme and Regularization Term}
	With the uncertainty map, we develop an uncertainty-wise weighting scheme for segmentation loss.
	The scheme assigns the high values to pixel predictions with high uncertainty.
	Then a larger gradient would flow through the pixels with high uncertainty in the backward propagation.
	
	Taking the cross-entropy loss as an example, 
	suppose $Y\in\mathbb{R}^{H\times W\times C}$ denotes the ground-true segmentation mask with one-hot encoding, then the weighted loss takes the following form,
	\begin{equation}\label{uncertainty_ce}
	\mathcal{L}_{ce}(P_s)=-\frac{1}{HW}\sum_{h,w,c}\text{detach}(M+1)Y\log P_s,
	\end{equation}
	where $\text{detach}(\cdot)$ stops the gradient from flowing in the backward direction.
	The gradient of Eq.~\ref{uncertainty_ce} can be formulated as,
	\begin{equation}\label{eq_gradient}
	\frac{\partial \mathcal{L}_{ce}}{\partial P_s}=-\frac{1}{HW}\sum_{h,w,c}\frac{M+1}{P_s}Y.
	\end{equation}
	Obviously, according to Eq.~\ref{eq_gradient}, the gradient is highly related to $M$.
	For high-uncertainty pixels with large values in $M$ will generate larger gradients to update the model, guiding the optimizer to pay more attention to the high-uncertainty regions.

	In addition, to further reduce the negative impact of prediction uncertainty, we adopt an uncertainty regularization term for the proposed method.
	This regularization item reduces the uncertainty of segmentation predictions from a global perspective, and it can be formulated as:
	\begin{equation}
	\begin{aligned}
	\mathcal{R}_{kl}=\frac{\sum_{h,w}M}{HW}=\frac{-\sum_{h,w,c}\left(P_s\log{P'} +P_a\log{P'}\right)}{HWC}.
	\end{aligned}
	\end{equation}
	In the training stage, the uncertainty regularization term continuously reduces the prediction uncertainty, improving the accuracy and reliability of the model.

	\subsection{Objective Function}
	
	In our model, there are two segmentation predictions from the segmentation head and auxiliary head, respectively.
	Both of them should be supervised by the ground truth segmentation mask.
	Besides, an uncertainty regularization term is adopted to reduce the model uncertainty.
	
	Considering that iris segmentation is a pixel-wise binary classification problem, we adopt the weighted version of the combination of cross-entropy loss and DiCE loss~\cite{milletari2016v} for segmentation.
	The final objective function can be written as
	\begin{equation}\label{final}
	\small
	\min\underbrace{\mathcal{L}_{ce}(P_s)+\mathcal{L}_{Dice}(P_s)}_{\text{segmentation head}}+
	\underbrace{\mathcal{L}_{ce}(P_a)+\mathcal{L}_{Dice}(P_a)}_{\text{auxiliary head}}+\alpha\mathcal{R}_{kl},
	\end{equation}
	where $\alpha$ is a trade-off parameter.
	$\mathcal{L}_{ce}$ is the weighted cross-entropy loss defined in Eq.~\ref{uncertainty_ce}, while $\mathcal{L}_{Dice}$ is the weighted Dice loss, i.e.,
	\begin{equation}\label{uncertainty_dice}
	\begin{aligned}
	\mathcal{L}_{Dice}(P_s)=1-\frac{1}{C}\sum_c\frac{2\sum_{h,w}M\times I+\epsilon}{\sum_{h,w}M\times U+\epsilon}
	\end{aligned}
	\end{equation}
	where $I=P_sY$ and $U=P_s+Y$ represent the soft intersection and soft union between prediction and segmentation mask, respectively.
	$\epsilon=1$ is the smoothing factor.

	
	\renewcommand\arraystretch{0.95}
	\begin{table*}[t]
		\centering
		\caption{Quantitative comparison of iris segmentation methods (\%).
			Note:
			1) We highlight the best results in bold and underline the second-best results.
			2) The state-of-the-art method and the baseline method are marked using $\dag$ and $\S$, respectively.}	
		\label{Table_seg}	
		\vspace{-3mm}
		\begin{tabular}{c|cc|cc|cc|cc|cc|cc}
			\hline
			\multirow{2}{*}{Methods} & \multicolumn{4}{c|}{CASIA-distance} & \multicolumn{4}{c|}{UBIRIS.v2} & \multicolumn{4}{c}{MICHE-1}  \\ \cline{2-13} 
			& E1$\downarrow$                 & F1$\uparrow$  & mIoU$\uparrow$ & Acc$\uparrow$   & E1$\downarrow$                 & F1$\uparrow$  & mIoU$\uparrow$ & Acc$\uparrow$  & E1$\downarrow$                 & F1$\uparrow$  & mIoU$\uparrow$ & Acc$\uparrow$  \\ \hline
			RTV-L1                    & 0.68   & 87.55   & 78.25 & 81.04 & 1.21  & 85.97  & 77.63      & 88.83      & 2.42 & 79.24 & 71.47  & 88.97      \\ \hline
			MFCNs                    & 0.59   & 93.09   & --   & --   & 0.90  & 91.04  & --   & --  & 0.74 & 92.01 & --  & --    \\
			CNNHT                    & 0.56   & 92.27   & 86.58   & 89.01  & 0.97  & 90.34  & 82.98 & 91.14 & 0.80 & 91.41 & 85.27 & 91.66  \\
			BiseNetv2                & 0.42   & 94.37   & 89.41   & 93.58  & 0.90  & 91.57  & 84.61 & 92.08 & 0.73 & 92.76 & 86.85 & 93.56 \\
			Unet                     & 0.42   & 93.96   & 88.84   & 91.28  & 0.91  & 91.59  & 84.67 & \underline{92.50} & 0.76 & 92.63 & 86.67 & 93.36 \\
			IrisParseNet $\dag$             & 0.41   & 94.25   & 89.52   & 93.29  & \textbf{0.84}  & {91.78}  & {84.88} & 92.31 & \underline{0.66} & {93.05} & {87.41} & {93.23} \\ 
			Swin-base$\S$                 & {0.40}   & {94.52}   & {89.68}   & \underline{93.91}  & 0.99  & 91.46  & 83.96 & 91.52 & 0.91 & 91.34 & 84.67 & 92.39 \\\hline
			BiTrans (ours)                    & \underline{0.38}   & \underline{94.84}   & \underline{90.23}   & {93.01}  & {0.87}  & \underline{91.83}  & \underline{85.07} & {91.49} & {0.68} & \underline{93.14} & \underline{87.48} & \underline{94.02} \\ 
			+ISUL (ours)                     & \textbf{0.35}   & \textbf{95.18}   & \textbf{90.84}   & \textbf{94.96}  & \underline{0.85}  & \textbf{92.11}  & \textbf{85.53} & \textbf{93.69} & \textbf{0.65} & \textbf{93.39} & \textbf{87.91} & \textbf{94.50} \\ \hline
		\end{tabular}
	\end{table*}
	
	\section{Experiment}

	\subsection{Databases}
	We evaluate the proposed approach on three publicly available datasets.
	These three datasets cover various application scenarios involving variations in illumination (NIR and VIS), sensors, distance, etc.
	
	\noindent{\textbf{CASIA-distance}}
	consists of 400 NIR images with a resolution of $640\times 480$, 300 images for training and 100 images for testing.
	
	\noindent{\textbf{UBIRIS.v2}} 
	is employed in the NICE-I competition.
	There are 500 VIS images for training and 445 VIS images for testing.
	The dataset adopts the resolution of $400\times300$.
	
	\noindent{\textbf{MICHE-I}} has 680 VIS images for training and 191 VIS images for testing.
	It has a resolution of $400\times400$.

	\subsection{Experimental Setting}
	
	In subsequent experiments, we implement the proposed approach on the platform of RedHat 4.8.5, Xeon(R) E5-2620, 1 TITAN X GPU, Pytorch 1.8.
	The experimental settings are detailed as follows.
	
	\noindent{\textbf{Baseline Methods}}
	involve seven iris segmentation approaches, including
	a one non-deep learning model
	(RTV-L1~\cite{zhao2015accurate}), 
	five CNN models 
	(MFCNs~\cite{liu2016accurate}, 
	CNNHT~\cite{hofbauer2019exploiting},
	BiseNetv2~\cite{yu2018bisenet, yu2020bisenet}, 
	U-Net~\cite{ronneberger2015u}, 
	IrisParseNet~\cite{wang2020towards})
	and one transformer model (Swin-base~\cite{liu2021Swin}).

	\noindent{\textbf{Evaluation Metrics}} come from existing work for iris segmentation and semantic segmentation, including
	E1 (the average segmentation error rate), 
	F1 (the harmonic mean of precision and recall), 
	IoU (intersection over union, the extent of overlap of the segmentation prediction and segmentation mask),
	Acc (the ratio of the correctly segmented region over the segmentation mask).
	Besides, we visualize the segmentation comparison map for qualitative comparison.
	In the segmentation comparison map, true positive (correct prediction), false positive (wrong prediction in the non-iris region), and true negative pixels (wrong prediction in the iris region) are marked in blue, red, and green, respectively.

	\noindent{\textbf{Data Augmentation}}
	is performed to enrich the data diversity, including\footnote{We used the image augmentation library Albumentations and adopted its default parameters.}
	(1) grid distortion with random distorting coefficient,
	(2) four-point perspective transform with random distances of the subimage's corners from the full image's corners
	(3) rotation at a random angle between -60 and 60.
	(4) spatial translation in x and y axis with random shift factor.

	\noindent{\textbf{Parameters}} are set as the following description.
	In the proposed method, we recommend $\alpha=0.001$ (Eq.~\ref{final}).
	As for the optimizer, we employ AdamW with a cyclical learning rate policy for 30000 iterations to train our model.
	Under the cyclic learning rate policy, the optimizer has a minimum learning rate of $10^{-5}$ and a maximum of $10^{-3}$.
	The cycle length is set to 4000.
	We adopt the batch size of 10.

	\subsection{Segmentation Comparison}
\begin{figure}[t]
		\centering
		\includegraphics[width=0.48\textwidth]{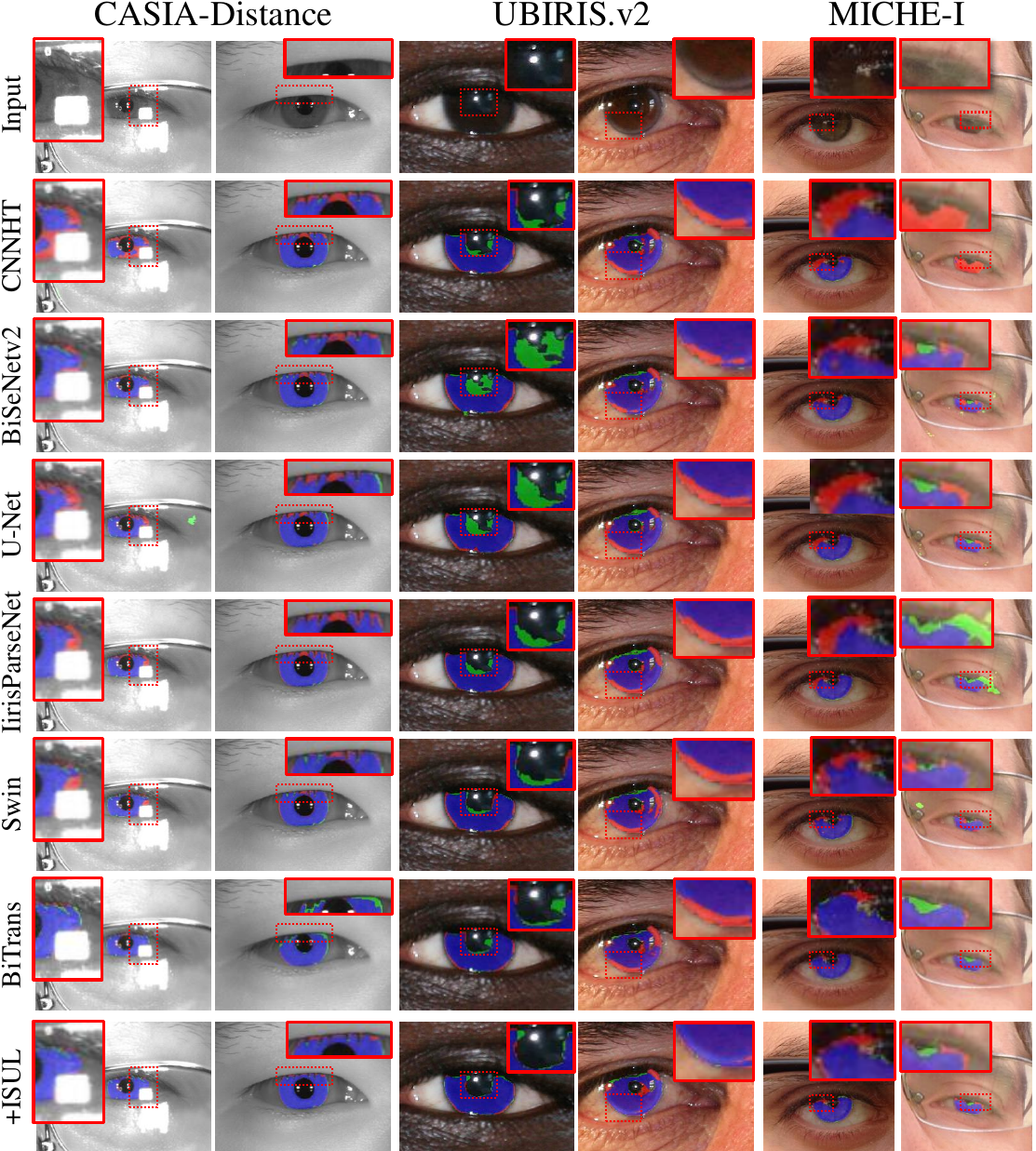}
		\vspace{-4mm}
		\caption{Qualitative segmentation comparison of different methods.
			Red boxes provide the close-up of the segmentation comparison map of different methods in each column.
			Appendix-C shows more qualitative comparison.		
			(Best viewed in color and zoom in)
		}
		\vspace{-6mm}
		\label{fig_comparison}
	\end{figure}

	This section conducts segmentation experiments on three datasets to evaluate the proposed approach.
	Table~\ref{Table_seg} lists summaries of the quantitative comparison, we report results from previous literature~\cite{wang2020towards} in `E1' and `F1' columns and supplement results in columns of `IoU' and `Acc' according to our implementation.
	Since we can not reproduce satisfying results for MFCNs, Table~\ref{Table_seg} does not show its results in terms of mIoU and Acc.
	The best and second-best results are highlighted in bold and underlined, respectively.
	Besides, Figure~\ref{fig_comparison} visualizes segmentation comparison maps.

	\noindent{\textbf{BiTrans}}
	surpasses the compared methods in most metrics (except E1) on multiple datasets.
	In terms of E1, BiTrans also shows competitive performance compared with the SOTA method (IrisParseNet).
	The superior performance of BiTrans mainly comes from two aspects.
	First, long-range contextual aggregation is a discriminative clue for accurate iris segmentation.
	The satisfying performance of Swin, the other transformer model, also supports our analysis.
	Second, the bilateral self-attention module learns spatial and visual information via two parallel branches for accurate segmentation.
	Since there are multiple ocular components with special spatial and visual characteristics in the acquired images, spatial and visual contextual information is crucial for iris segmentation.
	Both BiTrans and Swin-base adopt the shifted window strategy to learn spatial context information, yet our model utilizes an additional (visual) branch with a large receptive field to explore the visual relationship.
	Obviously, the visual branch discovers another discriminative clue, improving the segmentation performance on multiple datasets.
	Besides, BiTrans is elaborately designed for a better balance between performance and computational consumption.
	Compared to IrisParseNet, BiTrans achieves better performance using 35\% of Params and 20\% FLOPs.
	The leading advantages in performance and computing resources indicate the efficacy of the proposed method in iris segmentation.
	
	\noindent{\textbf{ISUL}} is adopted for weighting and regularization here, and the last row of Table~\ref{Table_seg} lists its segmentation results.
	According to the experimental results, applying uncertainty learning is beneficial to better segmentation performance.
	Obviously, the performance improvement is contributed by the modules related to ISUL, i.e., uncertainty-wise weighting scheme and uncertainty regularizer.
	The former encourages a large gradient flow through the pixels with high uncertainty in optimization.
	In other words, the weighting scheme makes the optimizer pay more attention to the high-uncertainty region.
	The uncertainty regularizer promotes the segmentation performance by reducing the predictive uncertainty.
	The effectiveness of the regularization term also suggests that the estimated uncertainty truly reflects the reliability of the model.
	In addition, Figure~\ref{fig_comparison} depicts that the introduction of ISUL significantly improves the segmentation performance of BiTrans in the limbic area.

	\begin{figure}[t]
		\centering
		\includegraphics[width=0.45\textwidth]{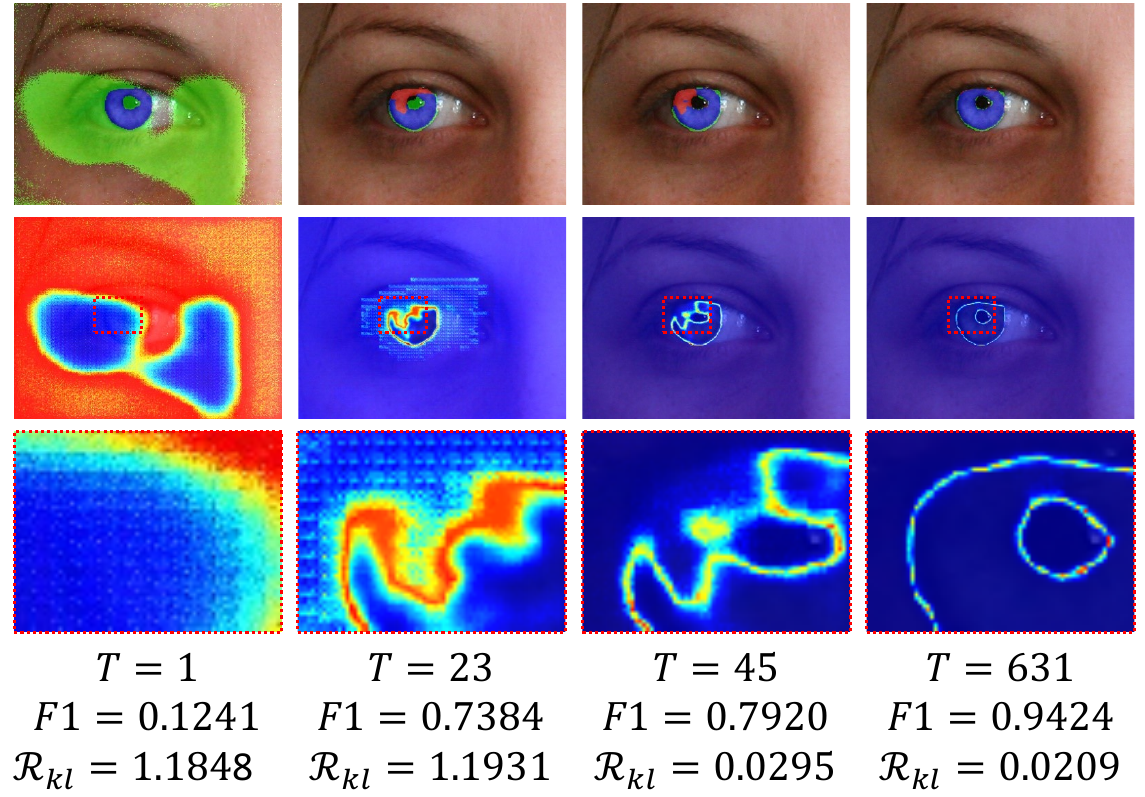}
		\vspace{-3mm}
		\caption{
			Segmentation comparison maps and uncertainty maps of a testing image at different training stages.
			Appendix-D provides more visualization results.
			(Best viewed in color and zoom in)
		}
		\vspace{-4mm}
		\label{fig_vis_train}
	\end{figure}

	\subsection{Visualization Analysis}
	
	Regarding the proposed algorithm, there is still a problem in our minds. 
	Does the uncertainty-wise weighting scheme really make the segmentation prediction more reliable? 
	And how does it work?
	To investigate this underlying mechanism, we provide a qualitative analysis by visualizing the uncertainty maps in this section.
	Figure~\ref{fig_vis_train} plots the uncertainty maps and segmentation comparison maps of a testing image at the different training stages.
	The uncertainty map can embody by Eq.~\ref{eq_reliable}.
	In Figure~\ref{fig_vis_train}, the first row shows the segmentation comparison maps.
	The second row visualizes the uncertainty maps using the heat map.
	The third row provides a close-up of the uncertainty maps.
	On the bottom row, we detail the information such as the number of epochs ($T$), F1, and the values of uncertainty regularization term.
	
	In the early training stage, the randomly initialized model has poor segmentation accuracy.
	Obviously, the current model could not provide a reliable segmentation prediction, i.e., pixels of the uncertainty map have very high uncertainty.
	As iteration increases, the model gradually has better segmentation performance. 
	The uncertainty map illustrates that high-uncertainty pixels are mainly located at the iris boundaries where the segmentation error occurs.
	In the subsequent optimization, high-uncertainty regions tend to produce higher loss values than other regions due to the weighting scheme.
	This weighted loss makes the model focuses on learning the segmentation knowledge in high-uncertainty regions.
	At the end of the iteration, segmentation errors rarely occur, but we can still find high-uncertainty regions in the limbic area.

	\subsection{Ablation Study}
	This section conducts the ablation study on the UBIRIS.v2 datasets to understand the contribution of each module.
	The ablation study involves six ablation conditions.
	\textbf{About BiTrans.}
	w/o TS: remove spatial branches.
	w/o TV: remove visual branches.
	w/o TC: discard convolutional projections and adopt linear projections.
	w/o TA: replace cross-attention mechanism using concatenation and self-attention mechanism.
	\textbf{About ISUL.}
	w/o UW: remove the uncertainty-wise weighting scheme.
	w/o UR: remove the uncertainty regularization term.
	Table~\ref{Table_ablation} lists the quantitative results of the ablation study, and the relevant experimental analysis is detailed in Appendix-B.


	\begin{table}[t]
		\centering
		\caption{Quantitative comparison for the ablation study on UBIRIS.v2 datasets (\%). 
		}
		\vspace{-3mm}
		\label{Table_ablation}
		\setlength{\tabcolsep}{3mm}{
			\begin{tabular}{c|cccc}
				\hline
				Methods & E1$\downarrow$                 & F1/DICE$\uparrow$  & mIoU$\uparrow$ & Acc$\uparrow$          \\ \hline
				Full  & \textbf{0.85}    & \textbf{92.11}  & \textbf{85.53} & \textbf{93.69}\\\hline
				w/o TS & {1.09}    & {89.64}  & {81.51} & {89.42}    \\ 
				w/o TV & {0.90}    & \underline{91.81}  & \underline{85.00} & \underline{92.91}    \\ 
				w/o TA & {0.91}    & {91.65}  & {84.74} & {92.26}    \\ 
				w/o TC & 0.95    & 91.50  & 84.29 & 92.03   \\ \hline
				w/o UW & 0.90    & 91.59  & 84.68 & {91.39}     \\
				w/o UR & \underline{0.89}    & 91.72  & 84.87 & {92.63}      \\\hline
				
		\end{tabular}}
		\vspace{-5mm}
	\end{table}
	\section{Conclusion and Limitations}
	The paper proposes a bilateral self-attention module and adopts it to design BiTrans, a transformer-style backbone for iris segmentation.
	Considering that the ocular components' spatial layout and visual characteristics are crucial clues for iris segmentation, the bilateral self-attention module adopts two branches to capture the information about spatial context without resolution reduction and visual context using a large receptive field, respectively.
	Besides, we develop ISUL by only introducing an auxiliary head for reliable segmentation.
	ISUL learns the uncertainty map according to the discrepancy between different predictions and applies the learned uncertainty to the weighting scheme and the regularization term.
	The experimental results on three iris datasets demonstrate the superior performance of the proposed approach, and visualization analysis investigates the underly mechanism of ISUL.

	Although promising results have been obtained in this paper, there are some open questions deserving further investigation.
	The most challenging one is iris segmentation in unconstrained scenes, especially for the low-quality iris images with low resolution.
	In the future, we will focus on reliable iris segmentation in unconstrained scenes.

	{\small
		\bibliographystyle{ieee_fullname}
		\bibliography{iris_seg}

\begin{thebibliography}{10}\itemsep=-1pt

\bibitem{antoran20depth}
Javier Antoran, James Allingham, and Jos\'{e}~Miguel Hern\'{a}ndez-Lobato.
\newblock Depth uncertainty in neural networks.
\newblock In H. Larochelle, M. Ranzato, R. Hadsell, M.~F. Balcan, and H. Lin,
  editors, {\em Advances in Neural Information Processing Systems}, volume~33,
  pages 10620--10634, 2020.

\bibitem{arsalan2019fred}
Muhammad Arsalan, Dong~Seop Kim, Min~Beom Lee, Muhammad Owais, and Kang~Ryoung
  Park.
\newblock Fred-net: Fully residual encoder--decoder network for accurate iris
  segmentation.
\newblock {\em Expert Systems with Applications}, 122:217--241, 2019.

\bibitem{banerjee2015iris}
Sandipan Banerjee and Domingo Mery.
\newblock Iris segmentation using geodesic active contours and grabcut.
\newblock In {\em Image and Video Technology}, pages 48--60, 2015.

\bibitem{bazrafkan2018end}
Shabab Bazrafkan, Shejin Thavalengal, and Peter Corcoran.
\newblock An end to end deep neural network for iris segmentation in
  unconstrained scenarios.
\newblock {\em Neural Networks}, 106:79--95, 2018.

\bibitem{bello2019attention}
Irwan Bello, Barret Zoph, Ashish Vaswani, Jonathon Shlens, and Quoc~V Le.
\newblock Attention augmented convolutional networks.
\newblock In {\em Proceedings of the IEEE/CVF international conference on
  computer vision}, pages 3286--3295, 2019.

\bibitem{cao2019gcnet}
Yue Cao, Jiarui Xu, Stephen Lin, Fangyun Wei, and Han Hu.
\newblock Gcnet: Non-local networks meet squeeze-excitation networks and
  beyond.
\newblock In {\em Proceedings of the IEEE/CVF International Conference on
  Computer Vision Workshops}, pages 1971--1980, 2019.

\bibitem{carion2020end}
Nicolas Carion, Francisco Massa, Gabriel Synnaeve, Nicolas Usunier, Alexander
  Kirillov, and Sergey Zagoruyko.
\newblock End-to-end object detection with transformers.
\newblock In {\em European Conference on Computer Vision}, pages 213--229,
  2020.

\bibitem{chen2021transunet}
Jieneng Chen, Yongyi Lu, Qihang Yu, Xiangde Luo, Ehsan Adeli, Yan Wang, Le Lu,
  Alan~L Yuille, and Yuyin Zhou.
\newblock Transunet: Transformers make strong encoders for medical image
  segmentation.
\newblock {\em arXiv preprint arXiv:2102.04306}, 2021.

\bibitem{chen2017deeplab}
Liang-Chieh Chen, George Papandreou, Iasonas Kokkinos, Kevin Murphy, and Alan~L
  Yuille.
\newblock Deeplab: Semantic image segmentation with deep convolutional nets,
  atrous convolution, and fully connected crfs.
\newblock {\em IEEE transactions on pattern analysis and machine intelligence},
  40(4):834--848, 2017.

\bibitem{daugman1993high}
John~G Daugman.
\newblock High confidence visual recognition of persons by a test of
  statistical independence.
\newblock {\em IEEE transactions on pattern analysis and machine intelligence},
  15(11):1148--1161, 1993.

\bibitem{devlin2018bert}
Jacob Devlin, Ming{-}Wei Chang, Kenton Lee, and Kristina Toutanova.
\newblock {BERT:} pre-training of deep bidirectional transformers for language
  understanding.
\newblock In {\em Proceedings of NAACL-HLT}, pages 4171--4186, 2019.

\bibitem{dosovitskiy2020image}
Alexey Dosovitskiy, Lucas Beyer, Alexander Kolesnikov, Dirk Weissenborn,
  Xiaohua Zhai, Thomas Unterthiner, Mostafa Dehghani, Matthias Minderer, Georg
  Heigold, Sylvain Gelly, Jakob Uszkoreit, and Neil Houlsby.
\newblock An image is worth 16x16 words: Transformers for image recognition at
  scale.
\newblock 2021.

\bibitem{he2008toward}
Zhaofeng He, Tieniu Tan, Zhenan Sun, and Xianchao Qiu.
\newblock Toward accurate and fast iris segmentation for iris biometrics.
\newblock {\em IEEE transactions on pattern analysis and machine intelligence},
  31(9):1670--1684, 2008.

\bibitem{hofbauer2019exploiting}
Heinz Hofbauer, Ehsaneddin Jalilian, and Andreas Uhl.
\newblock Exploiting superior cnn-based iris segmentation for better
  recognition accuracy.
\newblock {\em Pattern Recognition Letters}, 120:17--23, 2019.

\bibitem{jiang2021transgan}
Yifan Jiang, Shiyu Chang, and Zhangyang Wang.
\newblock Transgan: Two transformers can make one strong gan.
\newblock {\em arXiv preprint arXiv:2102.07074}, 2021.

\bibitem{lee2015deeply}
Chen-Yu Lee, Saining Xie, Patrick Gallagher, Zhengyou Zhang, and Zhuowen Tu.
\newblock Deeply-supervised nets.
\newblock In {\em Artificial intelligence and statistics}, pages 562--570,
  2015.

\bibitem{lian2018attention}
Sheng Lian, Zhiming Luo, Zhun Zhong, Xiang Lin, Songzhi Su, and Shaozi Li.
\newblock Attention guided u-net for accurate iris segmentation.
\newblock {\em Journal of Visual Communication and Image Representation},
  56:296--304, 2018.

\bibitem{liu2016accurate}
Nianfeng Liu, Haiqing Li, Man Zhang, Jing Liu, Zhenan Sun, and Tieniu Tan.
\newblock Accurate iris segmentation in non-cooperative environments using
  fully convolutional networks.
\newblock In {\em Proceedings of International Conference on Biometrics}, pages
  1--8, 2016.

\bibitem{liu2021Swin}
Ze Liu, Yutong Lin, Yue Cao, Han Hu, Yixuan Wei, Zheng Zhang, Stephen Lin, and
  Baining Guo.
\newblock Swin transformer: Hierarchical vision transformer using shifted
  windows.
\newblock In {\em Proceedings of the IEEE/CVF International Conference on
  Computer Vision}, pages 10012--10022, 2021.

\bibitem{lozej2018end}
Ju{\v{s}} Lozej, Bla{\v{z}} Meden, Vitomir Struc, and Peter Peer.
\newblock End-to-end iris segmentation using u-net.
\newblock In {\em Proceedings of IEEE International Work Conference on
  Bioinspired Intelligence}, pages 1--6, 2018.

\bibitem{milletari2016v}
Fausto Milletari, Nassir Navab, and Seyed-Ahmad Ahmadi.
\newblock V-net: Fully convolutional neural networks for volumetric medical
  image segmentation.
\newblock In {\em Proceedings of International Conference on 3D Vision}, pages
  565--571, 2016.

\bibitem{radford2019language}
Alec Radford, Jeffrey Wu, Rewon Child, David Luan, Dario Amodei, Ilya
  Sutskever, et~al.
\newblock Language models are unsupervised multitask learners.
\newblock {\em OpenAI blog}, 1(8):9, 2019.

\bibitem{ronneberger2015u}
Olaf Ronneberger, Philipp Fischer, and Thomas Brox.
\newblock U-net: Convolutional networks for biomedical image segmentation.
\newblock In {\em International Conference on Medical image computing and
  computer-assisted intervention}, pages 234--241. Springer, 2015.

\bibitem{shah2009iris}
Samir Shah and Arun Ross.
\newblock Iris segmentation using geodesic active contours.
\newblock {\em IEEE Transactions on Information Forensics and Security},
  4(4):824--836, 2009.

\bibitem{sun2019videobert}
Chen Sun, Austin Myers, Carl Vondrick, Kevin Murphy, and Cordelia Schmid.
\newblock Videobert: A joint model for video and language representation
  learning.
\newblock In {\em Proceedings of the IEEE/CVF International Conference on
  Computer Vision}, pages 7464--7473, 2019.

\bibitem{sun2008ordinal}
Zhenan Sun and Tieniu Tan.
\newblock Ordinal measures for iris recognition.
\newblock {\em IEEE Transactions on pattern analysis and machine intelligence},
  31(12):2211--2226, 2008.

\bibitem{sutra2012viterbi}
Guillaume Sutra, Sonia Garcia-Salicetti, and Bernadette Dorizzi.
\newblock The viterbi algorithm at different resolutions for enhanced iris
  segmentation.
\newblock In {\em Proceedings of International Conference on Biometrics}, pages
  310--316, 2012.

\bibitem{vaswani2017attention}
Ashish Vaswani, Noam Shazeer, Niki Parmar, Jakob Uszkoreit, Llion Jones,
  Aidan~N Gomez, {\L}ukasz Kaiser, and Illia Polosukhin.
\newblock Attention is all you need.
\newblock In {\em Advances in neural information processing systems}, pages
  5998--6008, 2017.

\bibitem{wang2020towards}
Caiyong Wang, Jawad Muhammad, Yunlong Wang, Zhaofeng He, and Zhenan Sun.
\newblock Towards complete and accurate iris segmentation using deep multi-task
  attention network for non-cooperative iris recognition.
\newblock {\em IEEE Transactions on information forensics and security},
  15:2944--2959, 2020.

\bibitem{wildes1997iris}
Richard~P Wildes.
\newblock Iris recognition: an emerging biometric technology.
\newblock {\em Proceedings of the IEEE}, 85(9):1348--1363, 1997.

\bibitem{yu2020bisenet}
Changqian Yu, Changxin Gao, Jingbo Wang, Gang Yu, Chunhua Shen, and Nong Sang.
\newblock Bisenet v2: Bilateral network with guided aggregation for real-time
  semantic segmentation.
\newblock {\em International Journal of Computer Vision}, pages 1--18, 2021.

\bibitem{yu2018bisenet}
Changqian Yu, Jingbo Wang, Chao Peng, Changxin Gao, Gang Yu, and Nong Sang.
\newblock Bisenet: Bilateral segmentation network for real-time semantic
  segmentation.
\newblock In {\em Proceedings of the European conference on computer vision},
  pages 325--341, 2018.

\bibitem{zaheer2020big}
Manzil Zaheer, Guru Guruganesh, Kumar~Avinava Dubey, Joshua Ainslie, Chris
  Alberti, Santiago Ontanon, Philip Pham, Anirudh Ravula, Qifan Wang, Li Yang,
  and Amr Ahmed.
\newblock Big bird: Transformers for longer sequences.
\newblock In {\em Advances in Neural Information Processing Systems},
  volume~33, pages 17283--17297, 2020.

\bibitem{zhao2015accurate}
Zijing Zhao and Ajay Kumar.
\newblock An accurate iris segmentation framework under relaxed imaging
  constraints using total variation model.
\newblock In {\em Proceedings of the IEEE/CVF International Conference on
  Computer Vision}, pages 3828--3836, 2015.

\bibitem{zheng2021rethinking}
Sixiao Zheng, Jiachen Lu, Hengshuang Zhao, Xiatian Zhu, Zekun Luo, Yabiao Wang,
  Yanwei Fu, Jianfeng Feng, Tao Xiang, Philip~HS Torr, et~al.
\newblock Rethinking semantic segmentation from a sequence-to-sequence
  perspective with transformers.
\newblock In {\em Proceedings of the IEEE/CVF Conference on Computer Vision and
  Pattern Recognition}, pages 6881--6890, 2021.

\end{thebibliography}
	}
	
\end{document}